\documentclass[conference]{IEEEtran}
\IEEEoverridecommandlockouts
\usepackage{cite}
\usepackage{amsmath,amssymb,amsfonts}
\usepackage{algorithmic}
\usepackage{graphicx}
\usepackage{textcomp}
\usepackage{xcolor}
\usepackage[bookmarks=false]{hyperref}       
\usepackage{url}            
\usepackage{float}

\DeclareMathOperator*{\argmax}{argmax} 

\ifCLASSOPTIONcompsoc
    \usepackage[caption=false,font=normalsize,labelfont=sf,textfont=sf]{subfig}
\else
    \usepackage[caption=false,font=footnotesize]{subfig}
\fi

\def\BibTeX{{\rm B\kern-.05em{\sc i\kern-.025em b}\kern-.08em
    T\kern-.1667em\lower.7ex\hbox{E}\kern-.125emX}}
\begin{document}

\title{Bidirectional Learning for Robust Neural Networks\\
\thanks{Partially funded by Norwegian Research Council under SOCRATES project (grant number 270961).}
}

\author{\IEEEauthorblockN{Sidney Pontes-Filho\IEEEauthorrefmark{1} and Marcus Liwicki\IEEEauthorrefmark{2}}
\IEEEauthorblockA{\IEEEauthorrefmark{1}\textsuperscript{,}\IEEEauthorrefmark{2}\textit{MindGarage, University of Kaiserslautern, Kaiserslautern, Germany}\\
\textit{\IEEEauthorrefmark{1}Department of Computer Science, Oslo Metropolitan University, Oslo, Norway}\\
\textit{\IEEEauthorrefmark{1}Department of Computer Science, Norwegian University of Science and Technology, Trondheim, Norway}\\
\textit{\IEEEauthorrefmark{2}Department of Computer Science, Electrical and Space Engineering, Lule{\aa} University of Technology, Lule\aa, Sweden}\\
Email: \IEEEauthorrefmark{1}sidneyp@oslomet.no,
\IEEEauthorrefmark{2}marcus.liwicki@ltu.se}}

\maketitle

\begin{abstract}
A multilayer perceptron can behave as a generative classifier by applying bidirectional learning (BL). It consists of training an undirected neural network to map input to output and vice-versa; therefore it can produce a classifier in one direction, and a generator in the opposite direction for the same data. The learning process of BL tries to reproduce the neuroplasticity stated in Hebbian theory using only backward propagation of errors. In this paper, two novel learning techniques are introduced which use BL for improving robustness to white noise static and adversarial examples. The first method is \textit{bidirectional propagation of errors}, which the error propagation occurs in backward and forward directions. Motivated by the fact that its generative model receives as input a constant vector per class, we introduce as a second method the \textit{hybrid adversarial networks} (HAN). Its generative model receives a random vector as input and its training is based on generative adversarial networks (GAN). To assess the performance of BL, we perform experiments using several architectures with fully and convolutional layers, with and without bias. Experimental results show that both methods improve robustness to white noise static and adversarial examples, and even increase accuracy, but have different behavior depending on the architecture and task, being more beneficial to use the one or the other. Nevertheless, HAN using a convolutional architecture with batch normalization presents outstanding robustness, reaching state-of-the-art accuracy on adversarial examples of hand-written digits.
\end{abstract}


\section{Introduction}
Deep neural networks present impressive performance in computer vision tasks, such as image classification and object detection \cite{imagenet, object_detection}, but they are vulnerable to small designed perturbations and visually unrecognizable images giving high confidence predictions \cite{dnn_easily_fooled,intriguing_prop_nn}. This vulnerability can cause severe security issues. Just imagine a self-driving car controlled by these neural networks. Are they reliable? In computer vision literature, there are attacking methods for crafting small perturbations that are imperceptible by the human eye, but result in deep neural networks incorrectly identifying them with absolute certainty \cite{intriguing_prop_nn}. The images produced by those attacking methods are called \textit{adversarial examples}. Other methods produce unrecognizable images for which deep neural networks give highly confident predictions \cite{dnn_easily_fooled}.

The idea of bidirectional learning (BL) is to make the output layer of a discriminative neural network only active when real input data is given, like the behavior of a generative classifier. That is done by teaching the same model to learn how to "read" (discriminative) and "write" (generative). Because of that, an undirected neural network can be a classifier and a generator at the same time and thereby improve classifier's robustness to random noise and adversarial examples. Our goal is to make multilayer perceptrons behave as a generative classifier, such as radial basis function \cite{rbf, rbf_universal}, deep Bayes classifier \cite{genclassifier}, and many others. Generative classifiers were identified as robust to adversarial examples \cite{genclassifier}. Weights and bias adaptation of multilayer perceptrons under BL is performed only by backward propagation of errors (backpropagation). Only real data is utilized for training the neural networks.

The main contribution of this paper is the introduction of two BL methods.\footnote{Complete project available at \url{https://github.com/sidneyp/bidirectional}.} The first method, called bidirectional propagation of errors, trains a hybrid undirected neural network to map images to labels (classifier) and labels to images (generator) in the opposite direction. The second method replaces the training of its generator by using the framework of generative adversarial networks (GAN) introduced by Goodfellow et al. \cite{gan}. This leads to hybrid adversarial networks (HAN), where the generator that has as input a latent variable and is trained by an adversarial discriminator. The HAN classifier uses the transposed weights of the generator. Therefore it contains a hybrid model which merges the generator and classifier. To evaluate the performance of these two approaches, we perform experiments on many models for measuring accuracy on unmodified test data, test data with noise addition, and adversarial test data. We also assess the robustness of the models to white noise static by checking their rates of maximum output for noise data over real test data.

\section{Related work}
\label{sec:rel_work}

Bidirectional learning has similarities to deep belief networks (DBNs) \cite{deepbeliefnet} because they are also hybrid models. However, DBNs perform a pre-training phase with restricted Boltzmann machines (RBM) \cite{deepbeliefnet, rbm} for an unsupervised input reconstruction layer-by-layer, from training data input layer to a final associative memory. Then an output layer for the discriminative model is added representing the ground truth, and backpropagation is executed for a fine-tuned classification training. Some autoencoder frameworks contain encoder and decoder sharing their weights for dimensionality reduction tasks. Such "mirrored" autoencoders are described in \cite{xu1993least,hinton2006reducing}. There exist also deep hybrid models \cite{deep_hybrid_models} where discriminative and generative models share the same latent variables. Another similar method, called Eigenboosting \cite{eigenboost}, which its authors present a generative classifier by its hybrid training with Harr-like features \cite{viola_jones}.

Since the discovery that deep neural networks for image classification can be easily fooled by random noise; unrecognizable images; and adversarial examples \cite{dnn_easily_fooled,intriguing_prop_nn}, several defensive and attacking strategies were described in literature \cite{towards_adv_evaluation,explain_adv,cleverhans}. One way to make neural networks more robust is adversarial re-training \cite{explain_adv,learning_with_adv,adv_atk_def}. It consists of generating adversarial examples every epoch or iteration, and using them as training data. Another defensive strategy is adding an auxiliary classifier for adversarial examples detection \cite{safetynet,detecting_adv,adv_atk_def}. Since the creation of adversarial examples is by adding noise into real data, a denoising method can be useful. Therefore \cite{towards_adv_robust} applies denoising autoencoder before feeding the data into a classifier. There are defensive methods that use generative adversarial networks. Adversarial perturbation elimination with GAN (APE-GAN) uses the generator of GAN as a denoising autoencoder \cite{adv_survey,ape_gan}. Another method using GAN is the Generative Adversarial Trainer \cite{adv_survey,gat} which the generator of GAN produces adversarial perturbations into the training set. These previous defensive methods use adversarial examples during training, so those neural networks can be biased to the method which designs the adversarial examples. 

The method of network distillation increases the robustness without the need for adversarial examples in training set \cite{net_distillation,adv_atk_def}. Its idea is to train a neural network to behave as another trained neural network. Instead of giving hard labels to a neural network, the \textit{temperature}-controlled softmax output of the trained neural network is given as ground truth. However, \cite{distillation_not_robust} verified that network distillation is still vulnerable to adversarial examples. A generative classifier presented as a defensive method to adversarial examples is called Gaussian process hybrid deep neural networks \cite{gpdnn,adv_atk_def}. The last layer of that robust convolutional neural network architecture consists of radial basis function kernels. Therefore it behaves as a generative classifier. The authors state that their deep architecture knows when it doesn't know. A biologically inspired defense against adversarial examples for deep neural networks is presented by \cite{adv_survey,bio_adv_protection}. Its principle is the creation of highly nonlinear neural networks which produces a saturated weight distribution found in the brain. All these three previous methods do not use adversarial examples during training and that is also our goal in this work.

\section{Bidirectional learning}
\label{bidirectional_learning}

Bidirectional learning produces a classifier and a generator in undirected neural network using backward propagation of errors in both directions. So each direction of this network has its own biases and the weights are shared. The idea is that the same positive weights of the last layer of a generator for producing white pixels can be the first layer of a classifier for identifying white pixels. Negative weights are similar regarding black pixels. Formally and over-simplified, any perceptron without bias that contains a weight vector $\mathbf{w} \in \{-1,1\}, \exists w=1$ and an input $\mathbf{x} \in \{0,1\}$ which its output $\mathbf{y}=f(\mathbf{w}\cdot \mathbf{x})$, where $f$ is the threshold activation function defined by 
\begin{equation}
\label{eq0}
f(a) =
  \begin{cases}
    0       & \quad \text{if } a \leq 0\\
    1  & \quad \text{if } a > 0
  \end{cases}
\end{equation}

The perfect activation input $\mathbf{\hat{x}}=\argmax_{\mathbf{x}} \mathbf{w}\cdot \mathbf{x}$ must have active inputs for positive weights and inactive inputs for negative weights, therefore $\mathbf{\hat{x}}=max(\mathbf{w},0)$. It shows $\mathbf{w}$ can also be adapted to be a contrast template of $\mathbf{\hat{x}}$, so the perceptron becomes a generative classifier with that fast adaptation procedure by "copying" its input when an activation occurs, as the Hebbian theory states \cite{hebb-organization-of-behavior-1949}. In biology, a real neuron produces a back-propagating action potential where the activation that goes through the axon back-propagates to its dendrites for plasticity regulation \cite{Grewe2010}. When the activation output of $\hat{x}$ is back-propagated, the result is equal to itself and expressed by
\begin{equation}
\label{eq1}
\mathbf{\hat{x}}\equiv f(\mathbf{w}^T\cdot f(\mathbf{w}\cdot \mathbf{\hat{x}})).
\end{equation}

When different activation functions and multilayers are used, \ref{eq1} becomes an approximation. So bidirectional learning forces equivalence in these cases. We infer that adding a supporting backpropagation to a classifier in the opposite direction that it is normally used can make the classifier's outputs less active when non-real data are given as input and avoid the vulnerability to adversarial examples. Since biological neurons learn by inputs, bidirectional learning uses a common training algorithm of artificial neural networks for trying to mimic Hebbian learning to the excitatory synapses (positive weights), because "neurons wire together if they fire together" \cite{Lowel209}; and for anti-Hebbian learning \cite{Vogels2013} to the inhibitory ones, because negative weights are strengthened when classifier's inputs remain inactive.

\subsection{Bidirectional propagation of errors}
\label{bidirectional_prop}

\begin{figure}[!ht]
\includegraphics[width=0.4\textwidth]{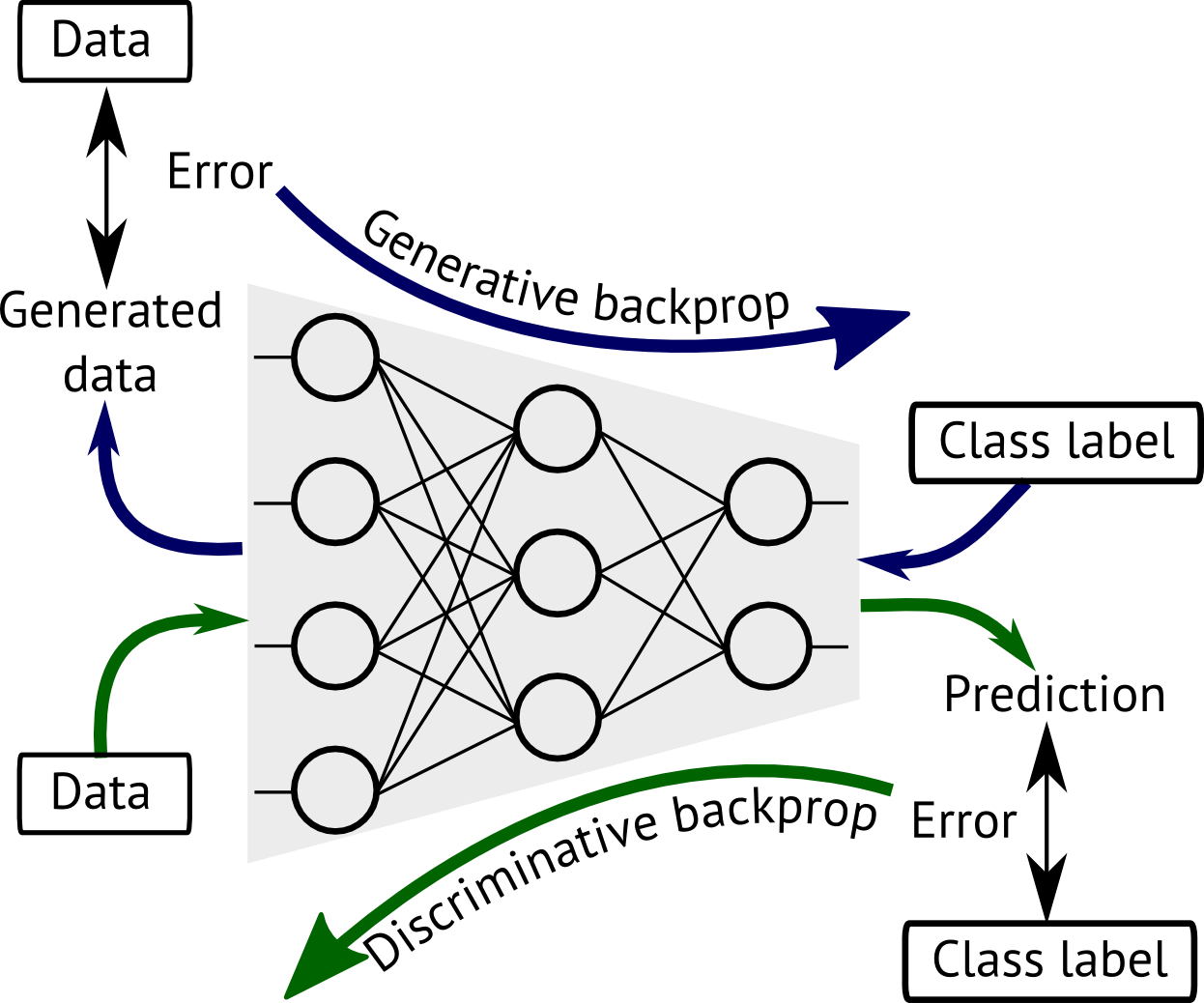}
\centering
\caption{Illustration of one training iteration in bidirectional propagation of errors. Dark green arrows represent the training with backpropagation (backprop) of discriminative model. Dark blue arrows represent the training of generative model. Same data and class labels are used for both in an iteration.}
\label{fig:biprop}
\end{figure}

Our supervised learning approach for both directions of a hybrid undirected neural network is the bidirectional propagation of errors. It consists of using backward propagation of errors (backpropagation) for mapping data to ground truth, and then ground truth to data. The mapping order can be reverted. The same batch of pairs of data and labels is utilized in normal and reversed backpropagation in a training iteration. Fig.~\ref{fig:biprop} shows how it works.

\subsection{Hybrid adversarial networks}
\label{ganc}

The previous method explained in Section~\ref{bidirectional_prop} has a limitation because the generator is trained with constant input per class. To avoid that, the hybrid adversarial networks (HAN) are introduced. There are three models in this framework based on GAN \cite{gan}: classifier $C$, generator $G$ that shares the same weights of $C$, and discriminator $D$ for being an adversary to $G$. The input of $G$ is a random vector $z$ of size equal to the number of classes for $C$. While $C$ is trained normally, $D$ and $G$ compete in a minimax game where $G$ tries to reproduce the real data to increase the error of $D$, and $D$ learns how to distinguish real data and data from $G$.

\begin{figure}[!ht]
\includegraphics[width=0.48\textwidth]{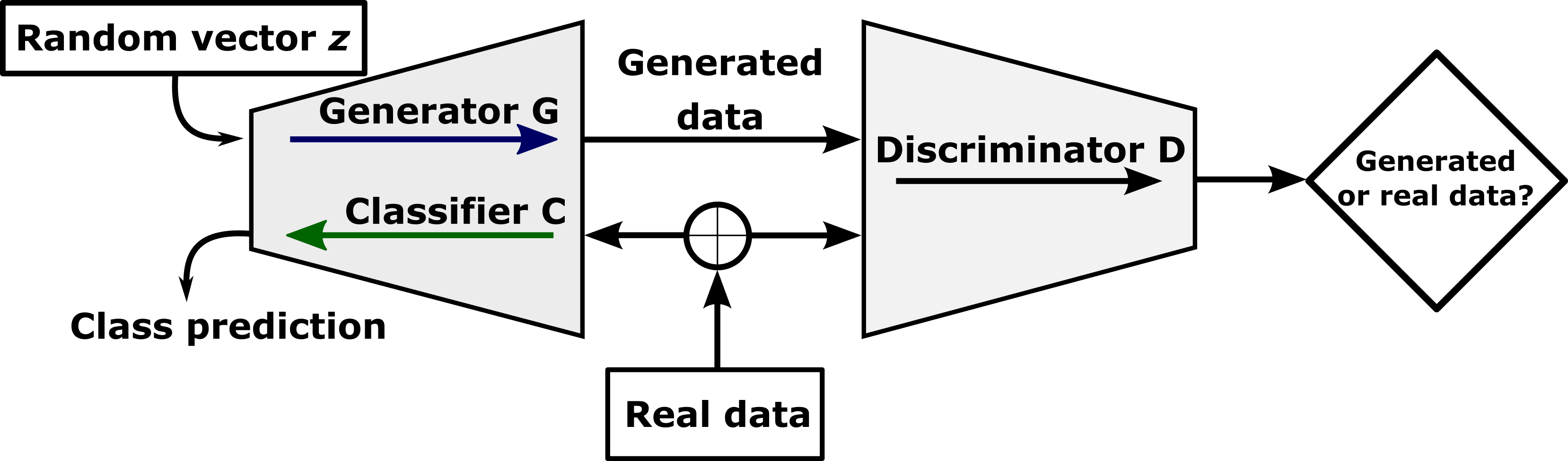}
\centering
\caption{Illustration of hybrid adversarial networks. Same color scheme is used in the hybrid model. Dark green arrow represents the discriminative model (classifier). Dark blue arrow represents the generative model.}
\label{fig:han}
\end{figure}

Our hybrid model merges $G$ and $C$, so the generator of GAN can be trained simultaneously as a transposed classifier for more robustness because we infer it can produce neurons that become active when images look "realistic". Fig.~\ref{fig:han} presents this framework. The training order in an iteration is $C$, then $D$, finishing with $G$.

\section{Experiments}
\label{experiments}

\begin{table*}[!t]
\centering
\caption{Description of architectures used on two methods with and without bias. NN stands for fully connected neural network and CNN for convolutional neural network. Convolutional layers are described by the number of kernels, inside the parenthesis is the kernel size and its stride (str).}
\label{architectures}
\begin{tabular}{|l|l|l|}
\hline
Method          & Architecture           & Units in discriminative hidden layer   \\ \hline
   & NN no hidden layer     & -                                              \\
Bidirectional    & NN one hidden layer    & 16         \\
propagation       & NN two hidden layers   & 16,16     \\
of errors       & NN four hidden layers  & 200,100,60,30               \\
                & CNN three conv. layers & 4 (5x5str1), 8(5x5str2),12(4x4str2),200    \\ \hline
Hybrid          & NN one hidden layer    & 128       \\
generative nets & CNN two conv. layers   & infoGAN architecture for MNIST \cite{infogan} \\ \hline
\end{tabular}
\end{table*}

These two methods were evaluated using the architectures with and without bias described in Table~\ref{architectures}. The architectures without bias are introduced to force all neurons in the network to have the same likelihood of activation and thereby increasing robustness. Architectures have been trained by mini-batch gradient descent with Adam optimizer \cite{adam} and mini-batch size of 100 data samples with ground truth (one mini-batch means one iteration). Bidirectional propagation of errors was trained with 50,000 iterations and HAN with 500,000 iterations because the adversarial training in HAN takes more time to converge. The implementation is based on TensorFlow 1.7 \cite{tensorflow}. The datasets used in the experiments are MNIST \cite{mnist} and CIFAR-10 \cite{cifar10}. Training set consists of 60,000 samples and test set of 10,000 samples. The adversarial attacking method to test the robustness of bidirectional learning is the fast gradient sign method (FGSM) \cite{explain_adv,cleverhans}. It disturbs real images to fool the classifier to make predictions for wrong classes. The equation for disturbing an image $x$ is 
\begin{equation}
\label{eq:fgsm}
x_{adv}=x+\epsilon* sign(\nabla_x J_\theta (x,y)),
\end{equation}

where the adversarial image $x_{adv}$ is produced by adding to the normal image $x$ with the sign method result of the gradient ascent $\nabla$ for $x$ of the loss function $J$ for model $\theta$ when image $x$ and label $y$ are given. This addition is limited by $\epsilon$ which is the maximum change in the pixels of $x$. The method which we use is from CleverHans v2.0.0 \cite{cleverhans}. The testing images were modified by FGSM with a max-norm epsilon ($\epsilon$) of 0.3 for MNIST, and 0.03 for CIFAR-10. Minimum and maximum pixel values of disturbed images are 0 and 1, respectively. We tested the robustness to white noise static by adding 10~\% of it into the test set for accuracy verification, and giving 100\% of that noise as classifier's input for measuring the sigmoid \cite{sigmoid} and softmax \cite{softmax} output layer. The maximum output for random noise $x_{noise}$ is divided by the maximum output for real test data $x_{test}$. Both $x_{noise}$ and $x_{test}$ have the same shape. That gives a rate of outputs to white noise static over real data. The sigmoid rate is for measuring output layer activity and expressed by
\begin{equation}
\label{eq3}
r_{sigmoid}=\frac{\max(C_{sigmoid}(x_{noise}))}{\max(C_{sigmoid}(x_{test}))}.
\end{equation}

The softmax rate for classification probability and formally denoted as
\begin{equation}
\label{eq4}
r_{softmax}=\frac{\max(C_{softmax}(x_{noise}))}{\max(C_{softmax}(x_{test}))}.
\end{equation}

All architectures with and without biases were trained by:
\begin{enumerate}
\item{Backpropagation (BP)}
\item{Bidirectional learning on first half of iterations, then backpropagation (BL then BP)}
\item{Bidirectional learning (BL)}
\end{enumerate}

\section{Results}
\label{results}

This section presents the results of our two methods on MNIST and CIFAR-10 dataset. It contains the accuracy for real test data, for test data with noise addition, and for test data modified by FGSM. The desired accuracy is 1.0 or 100~\%. We measure robustness to white noise static with sigmoid (activity) and softmax (class probability) rate of noise over real test data. The desired sigmoid output for noise data is 0.0 (fully inactive) and for test data is 1.0 (fully active). Therefore, the desired sigmoid rate is 0.0. Since we use datasets with ten classes, the desired softmax output for noise data is 0.1 or 10~\% confidence, and for test data is 1.0 or 100~\%. It means a softmax rate of 0.1. 

\subsection{Results of bidirectional propagation of errors}
\label{result_biprop}

\begin{figure*}[!t]
  \centering
    \subfloat[MNIST]{\includegraphics[width=\textwidth]{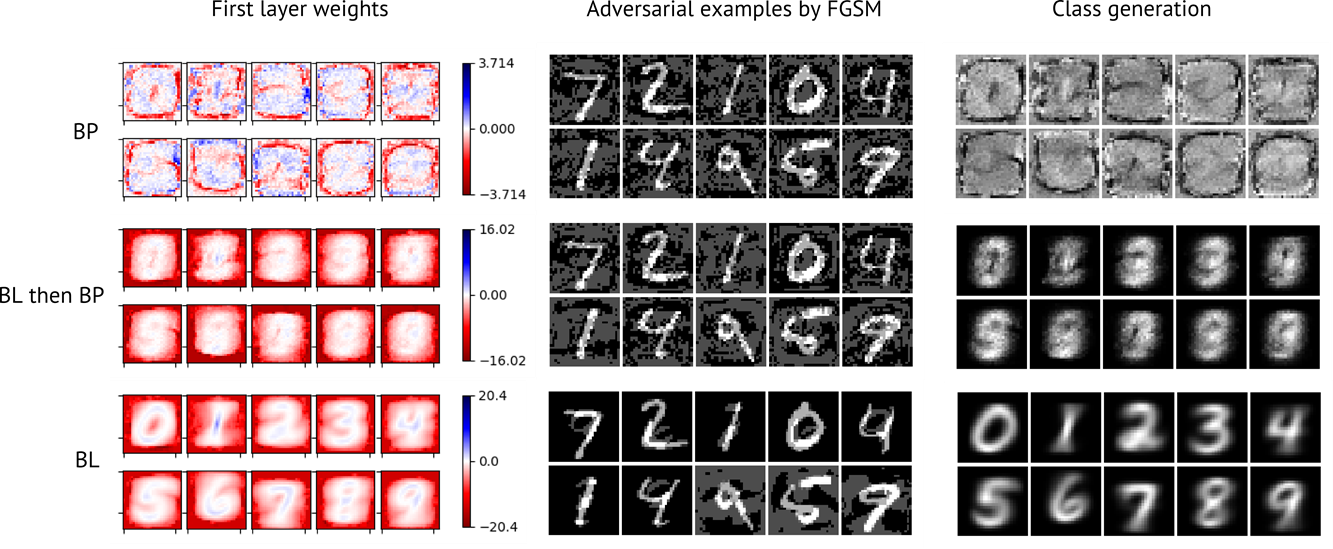}\label{fig:biprop_mnist}} \\   
    \subfloat[CIFAR-10]{\includegraphics[width=\textwidth]{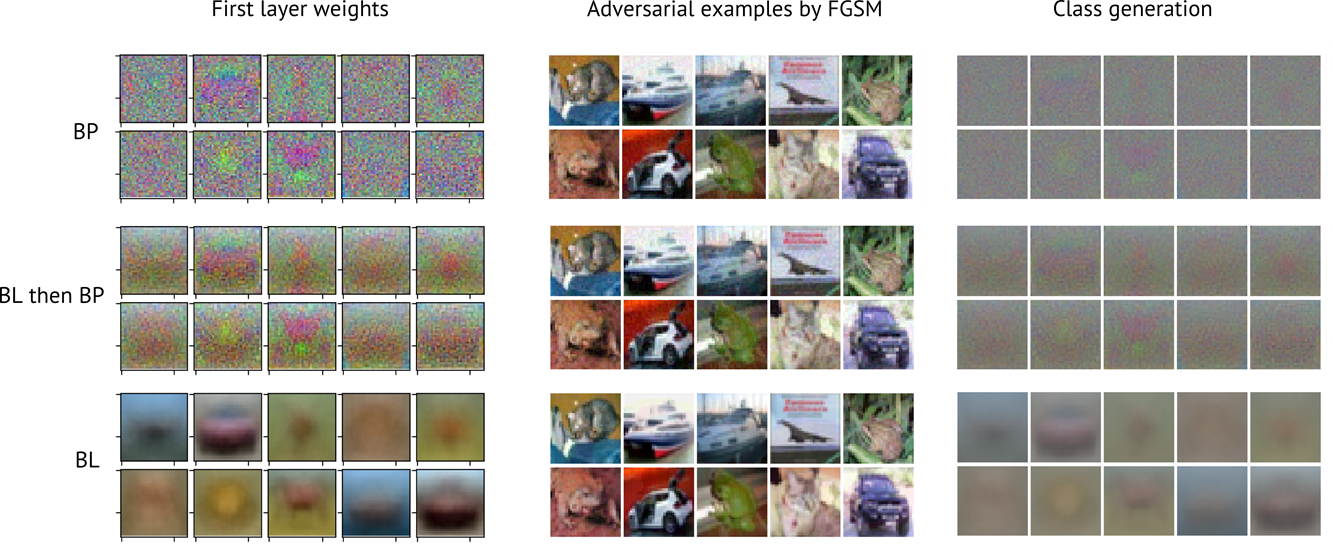}\label{fig:biprop_cifar}}    
    \caption[Short caption.] {\label{fig:res_biprop} Weights of the first layer, generated adversarial examples and images generated by a class label in bidirectional propagation of errors with a fully connected architecture without hidden layer and bias in all three learning methods on each row.}
\end{figure*}

Table~\ref{biprop_mnist_table} shows in the first row the architecture with most relative improvement in accuracy on adversarial examples. It is the architecture without hidden layer and bias, then it is a linear classifier. Backpropagation presents accuracy on adversarial examples of 4.17~\%, while bidirectional learning shows 60.14~\%. Since this is a simple architecture, the learned weights are easy to understand and to verify the causes of difference in robustness. Fig.~\ref{fig:biprop_mnist} shows that for MNIST dataset including adversarial examples and generated images for each class. The second row of Table~\ref{biprop_mnist_table} shows the best result regarding robustness to white noise static measured by the sigmoid and softmax rate of maximum output for noise over test data. The learning method that reached that was BL. The value of sigmoid rate is 0.5 which is a low value for sigmoid, meaning that the input for this activation function was zero. The value of softmax rate was the best value possible, 0.1 or 10~\%.

Table~\ref{biprop_cifar_table} shows the results of bidirectional propagation of errors for CIFAR-10 dataset. The first row contains the best relative accuracy improvement on adversarial examples. It is reached by the architecture trained by BL. Its weights, adversarial examples and generated images of all three learning methods are in Fig.~\ref{fig:biprop_cifar}. The order of CIFAR-10 classes is: airplane, automobile, bird, cat, deer, dog, frog, horse, ship, and truck. The weights of BP are noisy representations of those classes, but when we check the weights learned by BL, they are smooth and recognizable. We can see, for example, the weights of the blue color channel with high values for representing the sky and sea in airplane and ship classes. The second architecture in Table~\ref{biprop_cifar_table} presents an increase of 2.25~\% accuracy on test data when trained partially by BL compared with only BP. The architecture with two hidden layers and no bias is not shown here, but the training with BL then BP increases the accuracy on normal test data as well when compared to the results with BP.

\begin{table*}[!t]
\centering
\caption{Most significant result of bidirectional propagation of errors on MNIST. Selected iteration with best accuracy test. Bold numbers are the best results for each model.}
\label{biprop_mnist_table}
\begin{tabular}{|c|l|l|l|l|l|l|}
\hline
Model                         & Learning           & Accuracy & Accuracy  & Accuracy & Sigmoid           & Softmax  \\ 
 &    & test & noisy & adversarial &  rate          &  rate \\ \hline
Fully connected& BP    & \textbf{0.9273}        & \textbf{0.7138}         & 0.0417               & 3.34E-12        & 1            \\
no hidden                     & BL then BP & 0.9265        & 0.3216         & 0.045                & \textbf{0}                     & 1            \\ 
layer \& no bias                       &  BL      & 0.8781        & 0.6419         & \textbf{0.6014}               & \textbf{0}                     & 1            \\ \hline
Fully connected& BP    & \textbf{0.9456}        & \textbf{0.6502}         & 0.0318               & 0.9983          & 0.984   \\
one hidden                     & BL then BP & 0.9338        & 0.3807         & 0.06                 & 0.9923            & 0.6429 \\
layer \& no bias                       &  BL      & 0.905         & 0.5148         & \textbf{0.0814}               & \textbf{0.5}                   & \textbf{0.1}          \\ \hline
\end{tabular}
\end{table*}

\begin{table*}[!t]
\centering
\caption{Most significant result of bidirectional propagation of errors on CIFAR-10. Selected iteration with best accuracy test. Bold numbers are the best results for each model.}
\label{biprop_cifar_table}
\begin{tabular}{|c|l|l|l|l|l|l|}
\hline
Model             & Learning       & Accuracy & Accuracy & Accuracy    & Sigmoid      & Softmax      \\
                  &                & test     & noisy    & adversarial & rate          & rate        \\ \hline
Fully connected   & BP      & \textbf{0.3769}   & \textbf{0.373}    & 0.1853      & 0.9999   & 0.996 \\
no hidden         &BL then BP & 0.374    & 0.3678   & 0.1882      & \textbf{0}            & \textbf{0.9725} \\
layer \& no bias  &  BL      & 0.3211   & 0.3203   & \textbf{0.2711}      & \textbf{0}            & 0.9999    \\  \hline
Fully connected   & BP      & 0.4208   & 0.4137   & 0.351      & \textbf{0.9791} & 0.8627 \\
four hidden        & BL then BP & \textbf{0.4433}   & \textbf{0.4334}   & \textbf{0.3658}      & 0.9911 & 0.8359 \\
layers \& no bias  & BL      & 0.4314   & 0.4283   & 0.3596      & 0.9807 & \textbf{0.8289} \\ \hline
\end{tabular}
\end{table*}

\subsection{Results of hybrid adversarial networks}
\label{result_han}

\begin{figure*}[!t]
  \centering
    \subfloat[MNIST]{\includegraphics[width=0.5\textwidth]{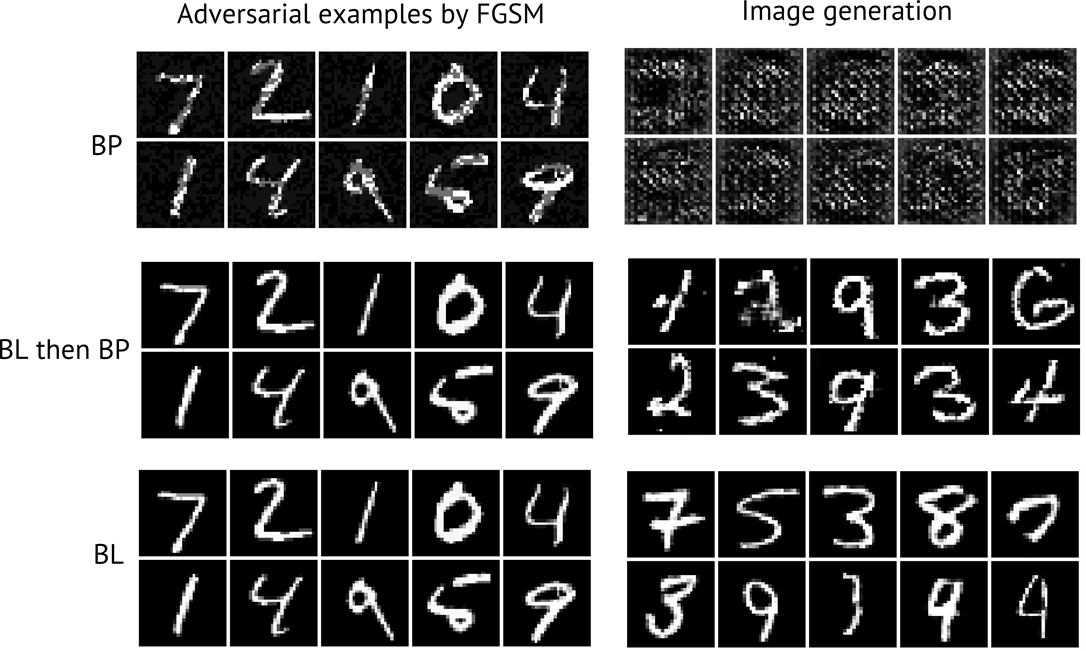}\label{fig:han_mnist}}
    \hspace*{\fill}
    \subfloat[CIFAR-10]{\includegraphics[width=0.5\textwidth]{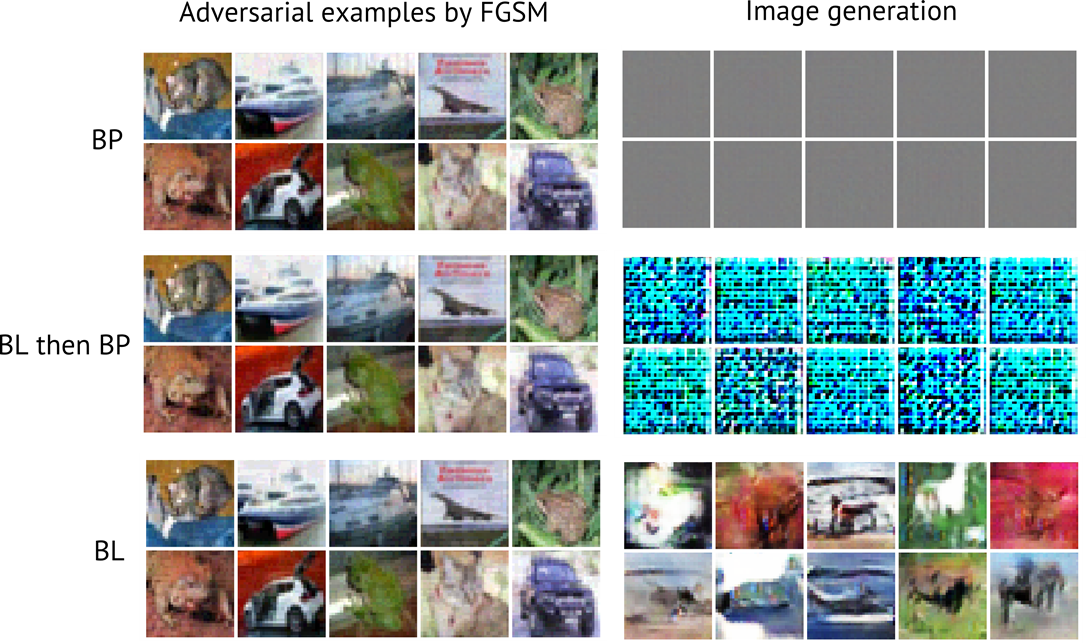}\label{fig:han_cifar}}
    \caption[Short caption.] {\label{fig:res_han} Generated adversarial examples and images generated by latent variable of hybrid adversarial networks in a CNN with two convolutional layers.}
\end{figure*}

Table~\ref{han_mnist_table} shows the best robustness of all experiments performed in this work. Hybrid adversarial networks on the architecture of infoGAN for MNIST \cite{infogan} achieved that. The model trained by BL then BP and without biases reached the 95.92~\% on adversarial examples of MNIST test set, while BP presented 5.08~\%. Even though there was a small reduction of accuracy on real test set compared with BP, from 99.21~\% to 98.49~\%. Fig.~\ref{fig:han_mnist} shows the adversarial examples for this architecture and the images generated by a random vector. That presents that HAN can also be a generative method when trained with BL. 

Table~\ref{han_cifar_table} is for the results of HAN on CIFAR-10 dataset. These results are not as good as the ones of HAN on MNIST dataset. The reason is that accuracy on real test set reduced drastically while presenting small improvement in robustness. However, Fig.~\ref{fig:res_han} presents that generator of HAN trained by BL has recovered the data distribution of CIFAR-10. Even though a generative model was not our goal.

\begin{table*}[!t]
\centering
\caption{Most significant result of hybrid adversarial networks on MNIST.  Selected iteration with best accuracy test. Bold numbers are the best results for each model.}
\label{han_mnist_table}
\begin{tabular}{|c|l|l|l|l|l|l|}
\hline
Model             & Learning       & Accuracy & Accuracy & Accuracy    & Sigmoid & Softmax   \\
                  &                & test     & noisy    & adversarial & rate    & rate      \\ \hline
CNN               & BP      & \textbf{0.9925}   & \textbf{0.9913}   & 0.0477      & 1       & 1         \\
two conv.         & BL then BP & 0.9854   & 0.9783   & \textbf{0.9375}      & 1       & 1         \\
layers            & BL      & 0.9823   & 0.9696   & 0.9084      & 1       & 1         \\ \hline
CNN               & BP      & \textbf{0.9921}   & \textbf{0.9906}   & 0.0508      & 1       & 1         \\
two conv.         & BL then BP & 0.9849   & 0.9768   & \textbf{0.9592}      & 1       & 1        \\
layers \& no bias & BL      & 0.9829   & 0.9491   & 0.9566      & 1       & 1         \\ \hline
\end{tabular}
\end{table*}

\begin{table*}[!t]
\centering
\caption{Most significant result of hybrid adversarial networks on CIFAR-10.  Selected iteration with best accuracy test. Bold numbers are the best results for each model.}
\label{han_cifar_table}
\begin{tabular}{|c|l|l|l|l|l|l|}
\hline
Model             & Learning       & Accuracy & Accuracy & Accuracy    & Sigmoid               & Softmax      \\
                  &                & test     & noisy    & adversarial & rate                  & rate         \\ \hline
CNN               & BP      & \textbf{0.7101}   & \textbf{0.6973}   & 0.161       & 1                     & 1            \\
two conv.         & BL then BP & 0.574    & 0.5645   & \textbf{0.258}       & 1                     & 1            \\
layers            & BL      & 0.565    & 0.5429   & 0.2445      & 1                     & 1            \\ \hline
CNN               & BP      & \textbf{0.7134}   & \textbf{0.7067}   & 0.1733      & 1                     & 1            \\
two conv.         & BL then BP & 0.5419   & 0.531    & \textbf{0.3366}      & 1                     & 1           \\
layers \& no bias & BL      & 0.4264   & 0.4114   & 0.1981      & \textbf{3.61E-07} & \textbf{0.9014}    \\ \hline
\end{tabular}
\end{table*}

\section{Analysis and discussion}
\label{analysis}

We can see in Fig.~\ref{fig:biprop_mnist} that backpropagation in the architecture without layer and no bias presents a noisy representation for each digit of MNIST dataset. Disturbances can be manually designed to this neural network; for example, the weights for number two have high positive values in the right. These positive weights in this part do not represent the most important white pixels in images with number two, but backpropagation identifies that part as the most relevant white pixels to recognize an image as number two. When increasing the pixel values in that part of an image with a different class, the resulting disturbed image can be recognized as number two. The adversarial examples generated by FGSM show that as well, and how noisy adversarial examples can be. Partial bidirectional learning (BL then BP) is similar to backpropagation, the only difference is that it knows that the pixels of the border are black because of high negative weights in that region. On the other hand, bidirectional learning performed in all iterations increased the robustness of this neural network. We can easily see the reasons in the learned weights, the adversarial examples, and the images generated by the label of each class. The creation of adversarial examples for BL became harder. FGSM tries in some of the test images to draw another number for fooling the neural network trained by BL.

By analysis of these characteristics, we infer the biological function of neural networks is not of a fine-tuned universal function approximator \cite{universal}, but it is of a multilayer contrast matching algorithm like a multilayer of Harr-like features from Viola-Jones object detection framework \cite{viola_jones}. The reason is that more robust weights present the characteristics of Harr-like features and that explains how real neurons learn so fast new complex patterns, just by "copying" the contrast of input that produces activation. The neurons of the primary visual cortex, from the retina through the lateral geniculate nucleus of the thalamus to V1 visual cortex \cite{neuroscience_book}, have the attributes for contrast detection of the weights trained by bidirectional learning or of the Harr-like features. The activation of a neuron depends on inputs with negative weights remaining inactive and inputs with positive weights being active. That also gives some light to the functionality of Hebbian and anti-Hebbian learning. The exclusion of bias makes neural networks more robust since it reduces the difference of neurons for activation likelihood. Therefore it tries to maintain neurons with equal importance in the network. Results on CIFAR-10 dataset show there are some architectures that when trained with full or partial BL can increase the accuracy on normal test data. Batch normalization also works to balance the neurons by keeping their inputs for activation function closer to zero. HAN results of infoGAN architecture without bias on MNIST dataset supports our analysis for equality in neuron importance and that a hybrid undirected neural network can be robust to adversarial examples.

\section{Conclusion and future work}
\label{conclusion}

Bidirectional learning produces a classifier and a generator in an undirected neural network, giving benefits to the classification task which is our main goal; moreover, it can also support generation of images too. Producing supporting methods and alternatives to backpropagation algorithm regarding robustness is essential for a reliable neural network. The defensive and learning method proposed in this paper was created by only adding a generative backpropagation in a discriminative multilayer perceptron. However, the difference of results on MNIST and CIFAR-10 dataset and on different architectures should be investigated.

For future work, we list the following advances possible after the proposal of bidirectional learning:

\begin{itemize}
\item{application of BL on different datasets and architectures;}
\item{the generator of bidirectional propagation of errors receiving as an input the label and the image together, then giving some variation to the generator's input and because of that it becomes an autoencoder;}
\item{hybrid adversarial networks framework can be improved like its first version for generation \cite{gan} because several improvements to GAN appeared since its introduction and they can be applied to extend HAN as well, but for classification purposes;}
\item{the decoder (generator) of an autoencoder as a transposed classifier;}
\item{weight decay can improve accuracy for data with white noise static and mimic non-Hebbian learning for positive weights and Hebbian learning for negative weights, because they can reduce weight of connections with, respectively, constantly active and inactive inputs, since constant inputs are meaningless for neurons like random inputs;}
\item{HAN can be verified as a generative method;}
\item{other tasks can be performed by coding input or desired output as images or binary strings;}
\item{alternatives to backward propagation of errors can be verified by analysis of BL behavior.}
\end{itemize}

\section*{Acknowledgments}

We are grateful for the support of Stefano Nichele in the review process and for the suggestions of all reviewers. We thank Benjamin Grewe for helpful discussions. We also thank Grant Sanderson of 3Blue1Brown channel on Youtube. His videos helped the main author of this work to come with bidirectional learning idea. Finally, we would like to thank Insiders Technologies GmbH for providing us with the necessary computational resources.

\bibliographystyle{IEEEtran}
\bibliography{IEEEabrv,bibexample}

\end{document}